\PassOptionsToPackage{numbers,sort&compress}{natbib}
\documentclass{article}

\usepackage[preprint]{neurips_2026}

\usepackage[utf8]{inputenc}
\usepackage[T1]{fontenc}
\usepackage{hyperref}
\usepackage{url}
\usepackage{booktabs}
\usepackage{amsfonts}
\usepackage{amsmath,amssymb}
\usepackage{nicefrac}
\usepackage{microtype}
\usepackage{xcolor}
\usepackage{graphicx}
\usepackage{enumitem}
\usepackage{listings}
\usepackage{subcaption}

\graphicspath{{images/}}

\newcommand{\minipic}{MiniPIC}
\newcommand{\ssep}{\textsc{SSep}}
\newcommand{\pdep}{\textsc{PDep}}

\title{MiniPIC: Flexible Position-Independent Caching in <100LOC\thanks{Code available at \url{https://github.com/IBM/vllm}, from commit \href{https://github.com/IBM/vllm/commit/6631ff34489612db6823b1ae3369c83eb6241853}{\texttt{6631ff3}} onwards}}

\author{%
  Nathan Ordonez\thanks{Main author.} \\
  IBM Research \\
  Zurich, Switzerland \\
  \texttt{nathan.or\-donez@ibm.com} \\
  \And
  Thomas Parnell \\
  IBM Research \\
  Zurich, Switzerland \\
  \texttt{TPA@zurich.ibm.com}
}

\begin{document}
\maketitle

\begin{abstract}

Retrieval-augmented and agentic workloads repeatedly prefill recurring predictable structured inputs (which we call 'spans') such as documents and code files. Yet, prefix caching in engines such as vLLM cannot reuse their KV entries unless they share identical prefixes with another request, while Position-Independent Caching (PIC) implementations within production-grade inference servers typically either require substantial server code changes or keep KV state outside the server, incurring host-to-device transfer overhead. We present Minimalistic PIC (MiniPIC): a minimal, flexible and fast vLLM design built from two ingredients: positional-encoding-free KV cache and user-controlled cache-reuse primitives. MiniPIC stores unrotated K vectors in the KV cache, applies RoPE to K tiles inside attention using per-request logical positions, and exposes three user-facing and token-level primitives: block-aligned padding, \ssep{}, and \pdep{}, that modify hashing behavior and effective block-level causal attention structure. With fewer than 100 lines of core-engine changes plus a custom attention backend, these primitives are sufficient to realize multiple PIC methods, including Block-Attention, EPIC, and Prompt Cache, within the same running vLLM instance, while natively integrating with KV cache CPU offload implementations. On 2WikiMultihopQA, MiniPIC with interleaved scheduling improves prefill throughput by 49\% over baseline vLLM, reduces cached-span time-to-first-token by up to two orders of magnitude, preserves the linear prefill scaling of uncached spans, and incurs only 5.7\% worst-case overhead.
\end{abstract}

\section{Introduction}
\label{s:intro}

Modern LLM inference is increasingly dominated by long-context prefill.
In retrieval-augmented generation (RAG), a query is expanded with multiple retrieved documents before generation~\citep{Lewis2020RAG}.
In agentic workloads, prompts repeatedly include tool descriptions, memory state, and execution traces~\citep{LinAgentInfer2025}.
These repeated segments recur across requests but not necessarily at the same absolute position or after the same prefix.

Production inference systems already exploit KV reuse through prefix caching.
vLLM~\citep{vllm2023}, followed by SGLang's radix-attention ideas~\citep{radixattention}, hashes fixed-size KV-cache blocks together with their preceding prefix.
This is powerful when many requests share the same initial prompt, but it misses an important locality pattern: two prompts may share the same document while differing in system prompt, document ordering, or surrounding context.
In those cases prefix caching treats the shared document as a cache miss.

A growing line of work proposes to prefill semantically independent prompt segments separately and later compose their KV states~\citep{PromptCache2023,BlockAttention2024,EPIC2024,CacheBlend2024,SuperpositionPrompting2024,REFRAG2025}.
We call such segments \emph{spans}.
Span techniques can dramatically increase cache hit rates, but implementing them inside a production inference server raises a systems question:
\begin{quote}
\emph{Should the KV cache contain positional encodings?}
\end{quote}

For RoPE-based models, standard inference stores keys after applying their positional rotation.
If a span prefilled at offset~0 is later reused at offset~2048, its cached K vectors carry the wrong rotation.
A single isolated request can be repaired by re-rotating the stored keys, but a serving system \emph{shares physical KV blocks across concurrent requests}.
If the same span is needed at two different offsets simultaneously, the same physical block cannot hold both rotations: this creates a fundamental concurrency conflict.

Recent work tackles PIC in vLLM-based serving through different approaches. MEPIC~\citep{MEPIC2025} identifies RoPE as a PIC barrier, stores position-free keys, fuses RoPE into attention, and introduces a chunk cache coordinator with its own eviction policy and a canonicalization module for page-aligned placement. SPNL~\citep{spnl} implements PIC through a procedural request interface that submits spans as individual prefills, but requires a full block repositioning and duplication engine with its CIDRA algorithm: a scheduler module that detects positional mismatches, generates block copy-or/and-reposition requests, and resolves cyclic memory movement dependencies with CPU scratch space.

Both prior implementations introduce significant new engine components. We instead frame flexible PIC as a \emph{separation-of-responsibilities} problem: which responsibilities belong inside the inference engine, and which can be left to the user or orchestration layer?
We argue that a flexible PIC implementation needs only two ingredients: (1) a positional-encoding-free KV cache inside the server, and (2) user-controlled cache-reuse mechanisms outside it.
The first removes shared-memory positional conflicts.
The second keeps the engine primitive small: rather than asking vLLM to pad chunks, recompute fixed interior blocks, or maintain dedicated chunk-cache state, we expose three user-facing primitives (padding, \ssep{}, and \pdep{}) that can be set by users or higher-level schedulers.
Our goals of performance, compatibility, and ease of deployment constrained us to find a solution that is easy to maintain while still supporting multiple span techniques.

\paragraph{Contributions.}
\begin{enumerate}[leftmargin=*,itemsep=2pt]
  \item \textbf{Systems analysis of positional-encoding conflicts.}
  We provide a detailed study of cascading problems (memory races, amplification, copy cycles, and accounting hazards) that arise when an inference server attempts PIC while retaining positional encodings in the KV cache (\S\ref{s:challenges}).

  \item \textbf{Minimal, flexible position-free KV implementation.}
  Our position-free KV implementation stores unrotated keys and applies RoPE in attention; excluding the modular attention backend, it requires only 78~LOC of vLLM core changes, of which only 61~LOC constitute new functionality (\S\ref{s:footprint}).

  \item \textbf{User-controlled cache reuse via three special tokens.}
  We show that three user-facing primitives (block-aligned padding, \ssep{}, and \pdep{}) are sufficient to express multiple span techniques while keeping vLLM simple. This gives control to the user or a higher-level scheduler/orchestrator rather than requiring engine-managed chunk canonicalization and mid-prompt recomputation policy (\S\ref{s:signaling}).

  \item \textbf{PIC-aware interleaved scheduling.}
  We introduce a scheduling policy that pipelines span prefill and final-prompt requests without synchronization barriers, achieving 49\% throughput improvement over baseline vLLM and 9\% over a barrier-separated baseline (\S\ref{s:scheduling}).
\end{enumerate}

\section{Background and Problem Statement}
\label{s:background}

\subsection{High-Performance LLM Inference and PIC}
\label{s:inference}

High-performance inference servers such as vLLM~\citep{vllm2023}, SGLang~\citep{radixattention}, and TensorRT-LLM~\citep{nvidia_tensorrt_llm} are optimized for throughput and latency in production deployments, distinguishing them from research-focused frameworks like HuggingFace Transformers~\citep{huggingface_transformers} or embedded engines like llama.cpp~\citep{llamacpp}.
The previously mentioned production systems (vLLM and SGLang) partition the KV cache into fixed-size blocks managed via paged attention~\citep{vllm2023}, with prefix caching assigning to each block a hash chained through the preceding block~\citep{radixattention}.
This enables cache hits only when both the block tokens \emph{and all preceding tokens} match exactly, enforcing correctness for ordinary causal attention but preventing reuse of repeated non-prefix spans.

A growing line of work proposes to prefill semantically independent prompt segments (spans) separately and later compose their KV states~\citep{PromptCache2023,BlockAttention2024,EPIC2024,CacheBlend2024,SuperpositionPrompting2024,REFRAG2025}.
Implementing span techniques inside a production inference server has led to diverse strategies with varying trade-offs.
Some approaches store spans outside the GPU accelerator, incurring host-to-device transfer overhead~\citep{CacheBlend2024,EPIC2024}.
Others implement PIC entirely within the inference engine but require extensive modifications to the scheduler, block memory mangement, and kernel execution pipeline~\citep{MEPIC2025,spnl}.
Many proposed PIC methods do not provide high-performance implementations suitable for production deployment~\citep{BlockAttention2024,SuperpositionPrompting2024,REFRAG2025}.
These complications raise a fundamental systems question: should the KV cache contain positional encodings?

Many of the latest decoder-only LLMs use Rotary Position Embeddings (RoPE)~\citep{su2024roformer}, which standard implementations apply \emph{before} the attention kernel, storing already-rotated keys in the KV cache.
Separating positional encodings from the KV cache by storing unrotated keys and applying RoPE inside the attention kernel has been proposed in recent architectures~\citep{DeepSeekV2_2024,MEPIC2025} and in SPNL~\citep{spnl}.
This separation solves a fundamental conflict with PIC: if the same span is reused at different positions simultaneously, the shared KV block cannot hold multiple positional encodings at once.
SPNL tackles this problem directly with its CIDRA (Cycle-Independent Duplicate Resolution Algorithm) method, which performs block-by-block copy and repositioning with graph-based cycle detection~\citep{spnl}.
This sophisticated mechanism is necessary when performing KV block duplication because memory copies imply that the source and target destinations may overlap, which makes them dependent upon the order in which a set of operations is applied.

\subsection{Properties of a PIC Implementation}
\label{s:properties}

Implementing position-independent caching (PIC) inside a production inference server requires two capabilities: (1) controlling the effective attention mask to specify which tokens share the same attention window, and (2) ensuring that cached KV entries for a span can be reused regardless of where that span appears in a later prompt.
Both requirements push against the design assumptions of engines that assume a single causal prefix structure and store keys with baked-in positional encodings.

\paragraph{The alignment principle.}
We argue that a flexible PIC implementation should be \emph{aligned} with the inference engine's architecture: it should work within the existing logical components of the codebase rather than adding new subsystems.
This mirrors the information-hiding criterion for modular decomposition, an influential software engineering principle~\citep{parnas1972criteria}: design decisions that are likely to change should be encapsulated in modules such that changes made to them are hidden from the rest of the system.
Specifically: by keeping the hash-function modification (special tokens in the block hasher) and position-free cache (on-the-fly RoPE in the attention kernel) as the only new engine components, padding, attention mask modification, and specific scheduling of span-prefill and span-loading requests are hidden from the engine and thus are entirely up to a layer above and outside of the inference engine.
We quantify the resulting footprint of the implementation, which we see as a sign of success in accomplishing this, in~\S\ref{s:footprint}.

\subsection{Why Positional KV Caches Are Brittle}
\label{s:challenges}

The conflict between position-dependent KV caches and PIC manifests when the same span must serve concurrent requests at different offsets.
SPNL addresses this positional-encoding conflict through its CIDRA (Cycle-Independent Duplicate Resolution Algorithm) method, which performs block-by-block copy and repositioning with graph-based cycle detection~\citep{spnl}.
However, CIDRA introduces significant overhead, for example from memory amplification (each concurrent user requires a separate copy) and copy cycles (overlapping source-destination ranges potentially requiring CPU scratch space).
Furthermore, CIDRA's repositioning operations are unfused and batched: all layers' KV blocks must be repositioned before attention can proceed, consuming memory bandwidth without interleaving with compute.
We claim these are not incidental bugs but structural consequences of storing position-dependent data in shared memory while allowing multi-position reuse. Such problems and complexity can be avoided by removing positional encodings from the shared cache entirely.

\paragraph{Problem statement.}
Achieving flexible PIC inside a production inference server thus requires an implementation that simultaneously (a) removes positional conflicts from the shared KV cache, (b) gives users control over the effective attention mask, and (c) keeps the engine changes small enough to be maintainable and aligned with the existing architecture: it should not introduce new scheduler request types, new memory management subsystems, or external caching services.
We call this property \emph{align\-ment}: the degree to which an implementation fits within the existing logical components of the inference codebase.

\section{Methods}
\label{s:methods}

We present \minipic{}, a minimal PIC substrate built from two ingredients: user-controlled cache-reuse primitives (special tokens that modify the engine's block-hashing behavior) and a positional-encoding-free KV cache (unrotated keys with RoPE applied inside the attention kernel). We first describe the special-token interface that gives users control over the effective attention mask, then the position-free cache that enables concurrent span reuse, and finally the request-submission and scheduling procedures used in our benchmarks.

\subsection{User-Controlled Cache Reuse with Special Tokens}
\label{s:signaling}

\subsubsection{Three User-Facing Primitives}

We intentionally keep the server-side contract minimal.
Rather than asking the inference engine to canonicalize and pad spans, recompute fixed middle blocks, or manage specialized chunk-cache state, we expose three user-facing primitives implemented as special tokens from the model's vocabulary (configured via environment variables):
\begin{itemize}[itemsep=1pt]
  \item \textbf{Padding}: the user pads spans to KV-block boundaries.
  \item \ssep{} (\emph{span separator}): detaches the block from its preceding hash chain.
  \item \pdep{} (\emph{prompt depend}): forces the block hash to depend on the \emph{entire} preceding prompt.
\end{itemize}
The latter two primitives modify the engine's block-hashing behavior \emph{only when they appear at the start of a KV block}:
The modified hash rule is:
\begin{equation}
\label{eq:hash}
h_i =
\begin{cases}
H(h_0,\, x_i)       & \text{if } x_i \text{ starts with } \ssep, \\
H(h_0,\, x_{\le i}) & \text{if } x_i \text{ starts with } \pdep, \\
H(h_{i-1},\, x_i)   & \text{otherwise,}
\end{cases}
\end{equation}
where $h_0$ is a global seed constant and $H$ is a cryptographic hash function.

Padding gives each span a canonical block-aligned layout by appending padding tokens to spans, so that future special tokens can reliably appear at the beginning of a block and trigger special hashing behavior.
\ssep{} makes a span independently cacheable regardless of prefix, enabling the position-independent attention masks required by Block-Attention~\citep{BlockAttention2024}, EPIC~\citep{EPIC2024}, and Prompt Cache~\citep{PromptCache2023}.
\pdep{} prevents false hits for suffix blocks (e.g., a query after several detached documents) whose KV vectors genuinely depend on all previous context.

\subsubsection{User Contract and Safety}

This approach extends the contract between the user and the inference server: incorrect padding or marker placement can cause false cache hits.
We intentionally keep validation outside the engine, placing the additional complexity of span technique management on the user (or a higher-level orchestration layer such as in SPNL~\citep{spnl}), which allows the inference server to keep its central focus on LLM inference.

\subsection{Position-Free KV Implementation}
\label{s:positionfree}

\subsubsection{Mechanism}

\minipic{} stores unrotated key vectors $\tilde{k}_j$ and value vectors $v_j$ in the KV cache.
Query vectors follow the standard path: they are rotated once by the model's \texttt{RotaryEmbedding} layer before entering attention.
The attention kernel receives per-request logical positions and applies RoPE to K tiles on-the-fly:
\begin{equation}
  \mathrm{score}(t,j) = \frac{(R_t\, q_t)^\top\; (R_{\pi(j)}\, \tilde{k}_j)}{\sqrt{d}},
\end{equation}
where $\pi(j)$ is the logical position of cached key~$j$ in the \emph{current} request.
The physical KV block no longer has a unique position; the same block can be referenced by many concurrent requests at different offsets.
Position exists only as per-request metadata consumed inside attention.

Conceptually, this moves positional encodings from shared HBM-resident KV memory to temporary SRAM/registers inside the attention computation.
Under a fixed span-attention mask and without counting concurrency, it is algebraically equivalent to re-rotating every cache hit, but without mutating or duplicating shared blocks.

\subsubsection{Kernel Modification}

Baseline FlashAttention receives pre-rotated $Q_{\text{rot}}, K_{\text{rot}}$ and computes $\text{softmax}(Q_{\text{rot}}K_{\text{rot}}^\top/\sqrt{d})\,V$.
\minipic{} leaves K unrotated in memory and extends the kernel:

\begin{center}
\small
\begin{minipage}{0.44\linewidth}
\begin{verbatim}
// Baseline Flash-Attention
for tile in K_tiles:
  // K already rotated
  scores += dot(Q_rot, K_rot_tile)
\end{verbatim}
\end{minipage}
\hfill
\begin{minipage}{0.50\linewidth}
\begin{verbatim}
// MiniPIC Flash-Attention
for tile in K_tiles:
  K_tile = load(tile)         // no RoPE
  c, s = cos_sin[pos[tile]]  // lookup
  K_rot = rotate(K_tile,c,s) // fused
  scores += dot(Q_rot, K_rot)
\end{verbatim}
\end{minipage}
\end{center}

\noindent The added work is a $\cos/\sin$ table lookup and two fused multiply-adds per K element, a small linear term within the $O(n^2)$ attention computation.
V vectors are unchanged.

\subsubsection{Native Offloading Compatibility}

Because cached blocks are position-free and PIC is managed through vLLM's native hashing functionality, \minipic{} composes naturally with KV movement across memory tiers.
vLLM's native CPU/storage offload path~\citep{vLLM2025KVCacheOffloading} can move unrotated span blocks between GPU and host without needing to track or adjust positions.
This contrasts with approaches that rely on external services (e.g., LMCache~\citep{LMCache2025}) which introduce additional host$\leftrightarrow$device transfer overhead.

\subsection{Expressing Span Techniques through the Interface}
\label{s:submission}

\subsubsection{Request-Submission Procedure for Block-Attention}

To reproduce a standard Block-Attention~\citep{BlockAttention2024} PIC scheme using \minipic{}, the user follows these steps for a RAG prompt with $N$ documents:

\begin{enumerate}[leftmargin=*,itemsep=1pt]
  \item Tokenize the system prompt, then pad to a block boundary (16 tokens).
  \item For each retrieved document, tokenize it, insert \ssep{} at the first token, then pad to a block boundary.
  \item Prefill each padded document and the system prompt as a separate request. Because \ssep{} resets the hash chain, each document's KV blocks are cached independently of the document's position in any later prompt.
  \item For the final prompt, construct the full token sequence: \texttt{[system][\ssep{},doc1][\ssep{},doc2]...[\pdep{},task]}. Insert \pdep{} before the task instruction so that the task's hash depends on the entire preceding context, preventing false cache hits.
  \item Submit the final prompt as a single request. The inference engine loads the independently cached document blocks from the KV cache; only the system prompt and task instruction are prefilled.
\end{enumerate}

When multiple concurrent requests reuse the same document at different prompt offsets, \minipic{}'s position-free KV cache serves all concurrent accesses from the same physical blocks, avoiding the duplication overhead described in \S\ref{s:challenges}.

\subsubsection{Other Span Techniques}

The same three primitives generalize to another PIC method, namely EPIC/LegoLink~\citep{EPIC2024}. To implement EPIC using our scheme, omit \ssep{} for the first $k$ tokens of each document so those tokens hash-depend on preceding context and are recomputed, while the remainder cache-hits via \ssep{}; with block-level granularity, $k$ rounds to a block multiple.

\subsection{PIC-Aware Interleaved Scheduling}
\label{s:scheduling}

Span caching is only useful if missing spans are computed before final requests need them.
Na\"ively inserting a barrier between span-prefill and final-prompt batches wastes GPU cycles.
We propose an \emph{interleaved} scheduling policy:

\begin{figure}[t]
\centering
\fbox{%
\begin{minipage}{0.92\linewidth}
\small
\textbf{Interleaved Span-Prefill Scheduling (ISPS)}\\[3pt]
\textbf{Input:} logical requests $\{r_{i,k}\}$ grouped into mini-batches indexed by $k$, each with spans $\{s_{i,j}\}$ and suffix~$q_i$, populating a final ordered list of requests $out$.
We do, per mini-batch:
\begin{enumerate}[leftmargin=*,itemsep=1pt]
  \item Tokenize all spans; prepend \ssep{} token to each, pad to block boundaries.
  \item Deduplicate: identify spans that occur multiple times, and evict duplicates.
  \item Append one prefill-only request per span to $out$
  \item Emit final-prompt requests (with \ssep{}/\pdep{} markers) for cache-hit composition.
\end{enumerate}
Once $out$ is populated, control concurrency via \texttt{max\_num\_seqs}: vLLM's FIFO scheduling guarantees that short span prefills complete before their dependent final prompts are dispatched. Then, submit $out$ as a single ordered batch of requests to vLLM.
\end{minipage}%
}
\caption{\minipic{} interleaved scheduling eliminates barriers and overlaps span prefill with ongoing inference.}
\label{fig:scheduler}
\end{figure}

\paragraph{Why interleaving helps.}
Under the barriered (\emph{batched}) strategy, the GPU idles between the span-prefill phase and the final-prompt phase.
Interleaving allows vLLM's continuous-batching engine to overlap short span prefills with decode steps from other in-flight requests, keeping the GPU saturated.
The concurrency cap (\texttt{max\_num\_seqs}) prevents final prompts from being scheduled before their dependencies are met.

\paragraph{Throughput analysis.}
Let $U$ be the number of unique spans, $N$ the total number of requests, $T_s$ the average span prefill time, and $T_f$ the full-prompt prefill time without PIC.
With cache-hit rate~$r$ (the fraction of spans served from cache):
\begin{itemize}[itemsep=1pt]
  \item \emph{Baseline}: cost $\propto N \cdot T_f$.
  \item \emph{Batched PIC}: $U \cdot T_s + N \cdot T_f(1{-}r)$ plus barrier idle time.
  \item \emph{ISPS}: same asymptotic cost but eliminates barrier overhead and enables overlap, effectively increasing $r$ by reducing span eviction under concurrency.
\end{itemize}
When span redundancy is high ($r \to 1$), total cost approaches $U \cdot T_s$, independent of corpus size~$N$.
Recent work on structured prompt languages demonstrates automated correctness enforcement for span-based scheduling~\citep{spnl}.

\section{Experiments}
\label{s:experiments}

\subsection{Setup}

All experiments use a single NVIDIA H100 80\,GB GPU (CUDA~13.2, PyTorch~2.9.1) on a node with dual Intel Xeon Platinum 8474C CPUs (96 cores) and 2\,TB DDR5 DRAM.
We evaluate using the \texttt{Tulu3-block-ft} model~\citep{BlockAttention2024}, the same model used in the Block-Attention study, which is architecturally identical to Llama-3-8B~\citep{metallama3} but fine-tuned for span-style RAG attention.
All configurations use the Triton attention backend~\citep{triton}, as our position-free RoPE implementation is written in Triton.

For throughput measurements we process 12{,}576 samples from the 2Wiki dataset~\citep{2wiki}; each sample contains exactly 10 retrieved documents of ${\sim}$100 tokens each, following the dataset construction methodology from the SPNL study~\citep{spnl}. Each sample is preceded by a system prompt and followed by a task instruction.
Since span techniques target prefill, we measure prefill-only throughput (disaggregated prefill scenario): a $2\times$ speedup corresponds to halving the required prefill GPUs.

We evaluate two baselines and two variants of our implementation.
The baselines are: (1) vanilla vLLM with prefix caching (using the Triton backend), and (2) SPNL with its CIDRA repositioning algorithm~\citep{spnl}.
We focus on PIC implementations that live inside the inference server, excluding approaches like EPIC~\citep{EPIC2024} and CacheBlend~\citep{CacheBlend2024} which manage KV vectors outside the GPU in CPU/host memory.
Such external caching approaches have fundamentally different computational characteristics due to host$\leftrightarrow$device transfer overhead, making direct comparison less informative.
We were unable to evaluate MEPIC~\citep{MEPIC2025} due to time limitations stemming from the complexity of their implementation and the lack of publicly available code at the time of writing.

Our implementation variants are: (1) \minipic{} with position-free KV active but no span signaling (worst-case overhead), and (2) \minipic{} batched with span prefills in one batch followed by a barrier and then final prompts.
Our full system, \minipic{} with ISPS (interleaved scheduling), is evaluated separately with varying \texttt{max\_num\_seqs} settings.
For baseline and PIC-disabled configurations, we use full-batch strategy where each sample is a request and all samples are submitted as a single batch at once.

\subsection{Performance}

\begin{table}[t]
  \centering
  \caption{End-to-end prefill throughput on 12{,}576 2Wiki samples. Batched and ISPS configurations use batch size 64. The first two rows use full-batch strategy where each sample is a request and all samples are submitted as a single batch.}
  \label{tab:throughput}
  \small
  \begin{tabular}{@{}llccr@{}}
    \toprule
    \textbf{Implementation} & \textbf{Strategy} & \textbf{\texttt{max\_num\_seqs}} & \textbf{Samples/s} & \textbf{Speedup} \\
    \midrule
    Baseline vLLM & full-batch & 1024 & 32.21 & 1.00$\times$ \\
    \minipic{} & PIC disabled & 1024 & 30.37 & 0.94$\times$ \\
    \midrule
    SPNL (CIDRA) & batched & 1024 & 36.14 & 1.12$\times$ \\
    \minipic{} & batched & 1024 & 44.07 & 1.37$\times$ \\
    \midrule
    \minipic{} & ISPS & 64 & 44.51 & 1.38$\times$ \\
    \textbf{\minipic{}} & \textbf{ISPS} & \textbf{1024} & \textbf{48.01} & \textbf{1.49$\times$} \\
    \minipic{} & ISPS & 2048 & 47.67 & 1.48$\times$ \\
    \bottomrule
  \end{tabular}
\end{table}

Table~\ref{tab:throughput} shows that \minipic{} with ISPS achieves 48.01 samples/s, a \textbf{49\% improvement} over baseline vLLM and a \textbf{33\% gain} over SPNL (36.14).
Key observations:
\begin{itemize}[itemsep=2pt,leftmargin=*]
  \item \emph{PIC disabled overhead}: only 5.7\% (30.37 vs.\ 32.21), confirming minimal RoPE-in-kernel cost.
  \item \emph{\minipic{} vs.\ SPNL}: \minipic{} significantly outperforms SPNL (44.07 vs.\ 36.14), validating that removing positional encodings eliminates the overhead of CIDRA's unfused copy and reposition operations.
  \item \emph{ISPS vs.\ batched}: eliminating the barrier yields a consistent additional gain (48.01 vs.\ 44.07).
  \item \emph{Scheduling sensitivity}: \texttt{max\_num\_seqs}=1024 is optimal; too few limits concurrency, too many causes memory pressure, where cached spans are more likely to be overwritten.
\end{itemize}

\subsection{Microbenchmarks}

\begin{figure}[t]
  \centering
  \includegraphics[width=\linewidth]{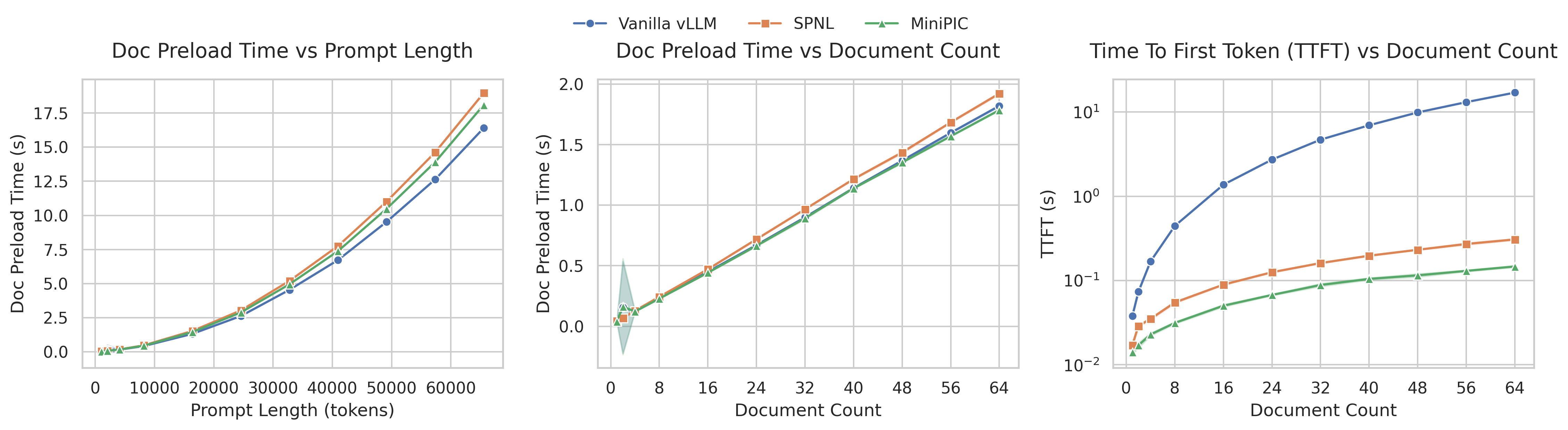}
  \caption{\textbf{Left}: Document preload time vs prompt length (no spans). \textbf{Middle}: Document preload time vs document count (1000-token synthetic documents prefilled in parallel). \textbf{Right}: Time-to-first-token (TTFT) vs document count, showing MiniPIC's significant improvement over SPNL and vanilla vLLM. The small differences between implementations in the left and middle plots indicate that worst-case slowdown is negligible, as reflected in Table~\ref{tab:throughput}'s PIC-disabled row.}
  \label{fig:microbenchmarks}
\end{figure}

Figure~\ref{fig:microbenchmarks} (left and middle) compares preload times across implementations.
The differences between vanilla vLLM, SPNL, and MiniPIC are small, indicating that the worst-case overhead of position-aware PIC is negligible.
This is consistent with the PIC-disabled row in Table~\ref{tab:throughput}, which shows only 5.7\% overhead for MiniPIC's position-free approach.

The right panel shows TTFT as cached documents are reused in a final prompt.
Vanilla vLLM suffers quadratic latency growth since it cannot reuse non-prefix documents.
SPNL achieves linear scaling through block duplication and repositioning, but MiniPIC significantly outperforms it (reflecting the throughput gains in Table~\ref{tab:throughput}) by eliminating CIDRA's memory-copy overhead entirely.

\subsection{Implementation Footprint and Alignment}
\label{s:footprint}

We now quantify the alignment of \minipic{} with vLLM's architecture by examining its implementation footprint. Table~\ref{tab:codechanges} categorizes every modified line of code as new functionality, routing, optional, or bug fix.

\begin{table}[t]
  \centering
  \caption{Source-code modifications for \minipic{}, categorized by change type. \emph{New functionality} implements the core algorithm; \emph{routing} passes existing objects without new logic; \emph{optional} can be reverted without affecting correctness; \emph{bug fix} addresses a pre-existing issue. Attention-backend changes are separated because vLLM supports modular backends.}
  \label{tab:codechanges}
  \small
  \begin{tabular}{@{}lccc@{}}
    \toprule
    File / Component & LOC & Category & Core? \\
    \midrule
    \texttt{envs.py} & 23 & New feature & Yes \\
    \texttt{rotary\_embedding/base.py} & 5 & New feature & Yes \\
    \texttt{core/kv\_cache\_utils.py} & 33 & New feature & Yes \\
    \texttt{core/block\_pool.py} & 4 & Bug fix & Yes \\
    \texttt{worker/gpu\_model\_runner.py} & 11 & Routing & Yes \\
    \texttt{platforms/cuda.py} & 2 & Optional & Yes \\
    \midrule
    \multicolumn{4}{l}{\textit{Attention backend (modular, interchangeable)}} \\
    \texttt{attention/backend.py} & 2 & Routing & No \\
    \texttt{attention/backends/triton\_attn.py} & 13 & Routing & No \\
    \texttt{attention/ops/triton\_unified\_attention.py} & 251 & New feature & No \\
    \midrule
    Total core & 78 & & \\
    ~~New core functionality & 61 & & \\
    Total (incl.\ backend) & 344 & & \\
    \bottomrule
  \end{tabular}
\end{table}

Outside the modular attention backend, \minipic{} requires 78~LOC of core changes, of which only 61~LOC constitute genuinely new functionality. The routing lines only pass information without introducing new logic. For example, in \texttt{gpu\_model\_runner.py} we simply pass the rotary embedding layer reference to attention metadata. The optional change (\texttt{cuda.py}) sets Triton as the default attention backend; and the 4~LOC bug fix in \texttt{block\_pool.py} prevents a double-free when the same block appears twice in one prompt, a pre-existing vulnerability exposed by span techniques.

For comparison, prior PIC implementations for vLLM introduce significantly larger engine changes.
SPNL~\citep{spnl} implements PIC through a procedural request interface with a full repositioning engine, including a scheduler module for block copying, graph-based cycle detection, and modifications to block allocation and reference-counting.
MEPIC~\citep{MEPIC2025} introduces canonicalization, a new hybrid KV manager with its own eviction policy, and scheduling modifications for ``prefill-in-the-middle'' operations, though their code is not publicly available at the time of writing.
Both implementations depart from the standard vLLM inference flow in multiple logical components, whereas \minipic{} touches only two: the block-hashing function and the attention kernel.

Note that the SPNL baseline in Table~\ref{tab:throughput} was built on vLLM~0.10.2, while \minipic{} uses vLLM~0.15.01.
While this version difference could introduce bias, our benchmarking shows negligible TTFT differences between these vLLM versions when running single requests of increasing size (as shown in the right panel of Figure~\ref{fig:microbenchmarks}).
The relative advantage of the position-free approach over CIDRA is therefore attributable to the algorithmic differences rather than version effects.
All our measurements use the Triton attention backend for fair comparison.

\section{Discussion and Related Work}
\label{s:discussion}

\paragraph{Design implications.}
Our experiments show that flexible PIC needs only two ingredients: position-free KV cache keys and user-controlled reuse mechanisms. Retaining positional encodings introduced the requirement of advanced memory management features (e.g., CIDRA's copy cycle detection). Pushing padding and reuse policy to the request/orchestration layer avoids engine-managed canonicalization and specialized caches. \minipic{} is therefore simpler (78~LOC core), faster (no copy/reposition overhead), and compatible with vLLM's CPU offload. ISPS contributes an additional $\sim$6\% throughput over batched submission by eliminating barriers. The same three primitives generalize to multi-turn conversations, agent frameworks, few-shot templates, or hierarchical documents.

\paragraph{Related work.}
Block-Attention~\citep{BlockAttention2024}, Prompt Cache~\citep{PromptCache2023}, EPIC~\citep{EPIC2024}, CacheBlend~\citep{CacheBlend2024}, Superposition Prompting~\citep{SuperpositionPrompting2024}, REFRAG~\citep{REFRAG2025}, and PICASO~\citep{PICASO2025} exploit independent span prefill with different composition strategies. Our work asks what inference-server primitives make such algorithms practical in vLLM. vLLM's PagedAttention~\citep{vllm2023} and SGLang's RadixAttention~\citep{radixattention} manage block-level caching; LMCache~\citep{LMCache2025} and Mooncake~\citep{mooncake} externalize KV storage for cross-instance sharing. Recent systems expose native offload APIs~\citep{vLLM2025KVCacheOffloading,Xie2025SGLangHiCache}. \minipic{} is complementary: it changes the representation of GPU-resident blocks so offload and reuse mechanisms no longer need position-specific copies. MEPIC~\citep{MEPIC2025} also proposes position-free keys, RoPE-in-attention fusion, and introduces a dedicated chunk cache; we take a smaller-surface approach, keeping vLLM's core changes minimal and moving responsibilities to a layer above vLLM. MEPIC's chunk-residency and eviction policies could thus be layered atop \minipic{}. StreamingLLM~\citep{xiao2023streamingllm} reapplies RoPE on-the-fly but without integration into production batching or caching. NoPE architectures~\citep{llama4} omit explicit positional encodings. MEPIC's code is not publicly available; our implementation is open-source at \url{https://github.com/IBM/vllm}.

\paragraph{Limitations and future work.}
\minipic{} targets RoPE-based transformers; extending to ALiBi or NoPE requires different handling. Incorrect marker placement can cause false cache hits, motivating pairing with a higher-level orchestration layer~\citep{spnl}. Span techniques approximate full causal attention, causing a reduction in accuracy which requires more future work. We keep vLLM's eviction policy unchanged; MEPIC's span-preserving two-queue policy~\citep{MEPIC2025} is complementary under high cache pressure. Public RAG traces with full timing and tokenized prompts are needed to evaluate online PIC benefits, and we were not able to find such traces despite our best efforts. \minipic{} does not guarantee a cached span remains in GPU memory until its dependent prompt is scheduled; under high concurrency, span eviction is possible, which MEPIC's chunk management or a cache-aware higher-level orchestrator as in SPNL could address. Integrating \minipic{} with sparse attention mechanisms~\citep{DeepSeekV32_2025} could combine linear-time prefill with effectively linear decode complexity, yielding end-to-end linear scaling for long-context inference.

\section{Conclusion}
\label{s:conclusion}

Span-based KV reuse can substantially reduce long-context prefill cost, but a flexible production implementation does not require a large engine-managed PIC stack.
We argue that two ingredients suffice: position-free keys in the KV cache, and a minimal user-controlled reuse interface built from padding, \ssep{}, and \pdep{}.
If shared KV blocks retain fixed positional encodings, in-place repositioning is incorrect under concurrency and CIDRA-style copy operations introduce memory amplification, copy cycles, and allocator complexity.
\minipic{} removes the root cause by storing unrotated K vectors and applying RoPE inside attention using per-request logical positions.
Combined with user-controlled reuse and PIC-aware interleaved scheduling, \minipic{} achieves 49\% higher prefill throughput on a large RAG workload and up to two orders-of-magnitude lower cached-span TTFT, with modest no-span overhead and 78 lines of core changes outside the modular attention backend.
For RoPE-based span-aware serving, we conclude that the simplest flexible PIC design is position-free in the cache and user-controlled at the interface boundary.

{
\small
\bibliographystyle{plainnat}
\bibliography{references}
}

\newpage
\appendix

\section{CIDRA: Analysis of Copy-Based Repositioning}
\label{apx:cidra}

When two requests share a span at different positions, SPNL's CIDRA algorithm allocates a fresh destination block, copies the source, and applies the RoPE delta.
This approach requires cycle detection and CPU scratch space for resolving overlapping copy dependencies.
CIDRA's throughput is limited because copy operations consume memory bandwidth without advancing generation, and its unfused, batched execution prevents interleaving with compute.
This confirms that copy-based repositioning introduces significant overhead compared to position-free approaches.

\section{Abandoned Design Alternatives}
\label{apx:abandoned}

\paragraph{Automatic padding and token replacement.}
An early prototype detected span markers anywhere in the prompt, padded spans inside the engine, and replaced markers with benign tokens.
Removed due to invasive frontend--scheduler state management.

\paragraph{Generalized $n$-back dependency tokens.}
A token grammar where a marker is followed by integer~$n$ specifying how many preceding tokens enter the hash.
Subsumes both \ssep{} and \pdep{}.
Rejected because recovering~$n$ requires detokenization during hashing.

\paragraph{Key-vector zeroing.}
Zeroing K vectors of special tokens would prevent the model from attending to markers.
Deferred due to attention-backend complexity.

\section{Hardware Details}
\label{apx:hardware}

\begin{table}[h]
  \centering
  \caption{Hardware platform for all reported experiments.}
  \small
  \begin{tabular}{ll}
    \toprule
    Component & Specification \\
    \midrule
    GPU & NVIDIA H100 80\,GB HBM3 ($\times$1 used) \\
    Host CPU & 2$\times$ Intel Xeon Platinum 8474C, 96 cores total \\
    Host DRAM & 2\,TB DDR5 \\
    Interconnect & NVLink full mesh (available, not used) \\
    CUDA & 13.2 \\
    PyTorch & 2.9.1 (CUDA backend) \\
    \bottomrule
  \end{tabular}
\end{table}

\end{document}